\documentclass[10pt,twocolumn]{article}
\usepackage[a4paper,top=1.8cm,bottom=1.9cm,left=1.7cm,right=1.7cm,columnsep=.7cm]{geometry}
\usepackage[T1]{fontenc}
\usepackage[utf8]{inputenc}
\usepackage{lmodern}
\usepackage{microtype}

\usepackage{graphicx,booktabs,tabularx,array,threeparttable}
\usepackage{amsmath,amssymb,mathtools,bm,amsthm}
\usepackage{xcolor,enumitem,caption,float,placeins,fancyhdr,csquotes,xurl}
\usepackage[backend=biber,style=numeric-comp,sorting=none,sortcites=true,maxbibnames=3,minbibnames=3,giveninits=true,doi=true,url=true,isbn=false]{biblatex}
\AtBeginBibliography{\sloppy}
\usepackage[colorlinks=true,allcolors=blue!55!black]{hyperref}
\graphicspath{{./}}
\setlist{nosep,leftmargin=*}
\newcommand{\E}{\mathbb E}
\newcommand{\Prob}{\mathbb P}

\newcommand{\eps}{\varepsilon}
\newcommand{\doop}{\operatorname{do}}
\newcommand{\KL}{D_{\mathrm{KL}}}

\newcolumntype{Y}{>{\raggedright\arraybackslash}X}
\newcolumntype{P}[1]{>{\raggedright\arraybackslash}p{#1}}
\newcommand{\PropositionName}{Proposition}
\newtheorem{proposition}{\PropositionName}
\newcommand{\runningtitle}{Observable-specific validity tests for scientific AI}
\fancypagestyle{plain}{\fancyhf{}\fancyfoot[C]{\thepage}}
\hypersetup{pdftitle={Observable-specific validity tests for scientific AI},pdfauthor={Xiaoshn Nee, Haobo Zhong, and Xiaomin Ni}}

\title{\textbf{Observable-Specific Validity Tests for Scientific AI}\\[.35em]
\large Controlled Approximations and Intervention Response}
\author{%
Xiaoshn Nee$^{a}$\thanks{Corresponding author. Email: \href{mailto:nixsh3@gmail.com}{nixsh3@gmail.com}}, Haobo Zhong$^{b}$, Xiaomin Ni$^{c}$\\[0.65em]
{\small $^{a}$Independent researcher}\\
{\small $^{b}$HSBC Business School, Peking University, Shenzhen City, Guangdong 518055, China}\\
{\small $^{c}$Artificial Intelligence Research Institute, Shenzhen University of Advanced Technology,}\\
{\small Shenzhen, China}}

\date{}
\begin{document}
\raggedbottom
\maketitle
\begin{abstract}
An approximation can reproduce an energy while missing its response, or satisfy an error tolerance without being certified by a given bound. We formulate observable-specific validity records for scientific artificial intelligence (AI) that link a reduction to its reference model, error measure, tolerance, parameter domain, and transformations or interventions. Explicit displacement and velocity certificates for a singularly perturbed oscillator recover distinct convergence orders and provide sufficient validity guarantees. Exact Hubbard dimer relations resolve the different coupling domains accepted by energy and response criteria. Across 16 repeated neural fits, imposing dimensional homogeneity reduces energy scale transfer error from 52.3\% to 0.0768\% and raises the median fraction jointly satisfying gap and response tolerances from zero to 100\%. Paired transformations verify the protected scaling relation, while a coupling crossover identifies the additional information needed for response transfer. A causal model recovers a positive intervention response beneath a strong negative observational association. For two mechanisms observed with Gaussian noise whose responses first differ at order $x^k$, we derive an exact fixed control query requirement scaling as $x^{-2k}$ and verify its full success curve numerically. Together with a 256-task evaluation interface, these results provide independently checkable tests of controlled approximation, physical transfer, and mechanism discrimination in scientific AI.
\end{abstract}
\par\medskip
\noindent\textbf{Keywords:} scientific AI; physical understanding; controlled approximation; effective models; causal inference; experimental design

\section{Introduction}
\label{sec:introduction}

A useful physical model often becomes more explanatory by retaining fewer details. In a strongly interacting electron system, virtual charge excitations can be replaced by a low energy spin exchange. In a strongly damped mechanical system, a rapidly relaxing momentum can be eliminated from the slow displacement dynamics. The resulting description succeeds when the retained variables and neglected processes have the right hierarchy for the observable being studied. Establishing this hierarchy gives numerical predictions their physical interpretation.

Scientific artificial intelligence (AI) increasingly supports this process through prediction, equation discovery, and the organization of scientific evidence\cite{carleo2019review,wang2023scientific}. Recent perspectives place large language models (LLMs) within the broader cycle of hypothesis formulation, modeling, and experimental assessment\cite{zhang2025method}. In physics, language models can execute multistep Hartree--Fock calculations when supplied with an appropriate decomposition\cite{hartreefock}. PhysReason, PHYBench, and the condensed matter benchmark CMPhysBench assess demanding physical problems\cite{physreason,phybench,cmphysbench}, while CritPt and PRL-Bench extend evaluation toward research calculations\cite{critpt,prlbench}. These capabilities motivate quantitative tests of the physical commitments that accompany a successful prediction.

The connection between prediction and understanding also has a substantial philosophical foundation. Contextual accounts relate understanding to knowledge that a scientific community can use and manipulate\cite{deregtdieks2005understanding}. Analyses of machine learning emphasize the evidential connection between a model and its target\cite{sullivan2022understanding,duede2023opacity}. We give this connection an operational form. A system demonstrates a specified physical capability when its explanation makes correct, testable commitments about reductions, limits, and interventions. This definition ties the evaluated capability to explicit scientific consequences.

Several recent studies directly motivate such tests. PhySense evaluates principle-based reasoning, including symmetry, dimensional analysis, power counting, and limiting cases\cite{physense}. PhysGym controls the prior information supplied during interactive discovery, NewtonBench changes physical laws and experimental environments, and LLM-SRBench tests equation discovery beyond common formula recall\cite{physgym,newtonbench,llmsrbench}. DiscoverPhysics separately evaluates predictive error and explanatory quality in noncanonical simulated worlds\cite{discoverphysics}. On the methodological side, KeplerAgent uses inferred physical structure to constrain symbolic regression, while explicit physical modeling has been separated from execution in physics reasoning\cite{kepleragent,modeling_execution}. Our focus is the \emph{validity of an approximation for each observable}. Reference quantities, acceptance tolerances, validity domains, and sufficient certificates become explicit evaluation objects. This adds an independently checkable validity decision to principle-based reasoning and interactive discovery.

Physical structure provides a productive basis for such commitments. Reviews of physics informed machine learning describe how equations, symmetries, and domain knowledge can guide learning\cite{karniadakis2021review}. Dimensional analysis constrains admissible functional forms\cite{buckingham,aifeynman}, and sparse discovery identifies compact dynamics within a chosen library\cite{sindy}. Hamiltonian, Lagrangian, and equivariant neural architectures encode conservation or transformation relations\cite{hnn,lnn,egnn}. Physics informed neural networks (PINNs) enforce differential equations through residual losses\cite{pinn}; studies of their optimization identify conditions under which these constraints are successfully learned\cite{pinnfail}. These approaches show the value of making a physical requirement explicit. Our framework extends this logic to the validity domain of a proposed reduction and to the observables it is intended to describe.

Our contribution is an evaluation record for individual observables linking a proposed reduction to its reference model, target quantities, validity domain, acceptance tolerance, and intervention consequences. We make the record executable through distinct tests for certified validity, measured error, transformation consistency, and joint observable acceptance. Explicit fast--slow certificates, exact dimer relations, paired neural experiments, and a known causal model instantiate these tests. An exact Gaussian measurement calculation then connects mechanism separation to query cost. The reported evidence consists of analytical models, neural surrogates, and controlled simulations; it supplies reference tests for subsequent language model evaluations.

\section{Observable-specific approximation certificates}
\label{sec:framework}

\subsection{The object being evaluated}

Let $\mathcal M(\bm p)$ denote a reference physical model with parameters $\bm p$. A proposed reduction $A$ specifies its retained variables or projection, effective equations, neglected terms, and controlling dimensionless quantities. We represent its claimed scope by
\begin{equation}
 \mathcal C=(\mathcal M,A,\mathcal Q,\mathcal P,\Omega,\bm s,\bm\delta,\bm b,\mathcal T),
 \label{eq:record}
\end{equation}
where $\mathcal Q$ is the set of target observables, $\mathcal P$ the parameter domain, $\Omega$ the observation window, $\bm s$ the normalization rules, $\bm\delta$ the acceptance tolerances, $\bm b$ the proposed error bounds, and $\mathcal T$ the specified transformations or interventions. For observable $Q$, define
\begin{equation}
 e_Q(\bm p;\Omega)=\frac{\|Q_A(\bm p)-Q_{\mathcal M}(\bm p)\|_\Omega}{s_Q(\bm p)}.
 \label{eq:error}
\end{equation}
The positive scale function $s_Q(\bm p)$ is declared before evaluation. The rule is fixed; its numerical value may depend on the parameters. Here displacement and dimensionless velocity use unit reference scales, gap error uses $s_J=J_{\mathcal M}(\bm p)>0$, and response-exponent error uses $s_n=1$. An intervention difference uses a declared nonzero response scale; the dimensionless causal example below uses $s_{\Delta Y}=1$. A certificate establishes $e_Q\le b_Q$ by analysis or a justified bound. For domain bookkeeping set $b_Q=+\infty$ outside its proof scope, while retaining an explicit out-of-scope label. Its certified domain satisfies
\begin{align}
 D_Q^{\rm cert}&=\{\bm p\in\mathcal P:b_Q(\bm p)\le\delta_Q\}\nonumber\\
 &\subseteq D_Q^{\rm valid}
 =\{\bm p\in\mathcal P:e_Q(\bm p)\le\delta_Q\}.
 \label{eq:domains}
\end{align}
Equation~\eqref{eq:domains} gives two complementary outputs. The certified domain provides a sufficient guarantee, while the reference error determines actual acceptance wherever it can be evaluated. Points with $b_Q>\delta_Q$ remain eligible for reference based verification. Physical tolerances such as $\delta_J=0.05$ describe the approximation itself; the numerical answer matching tolerance used by the evaluation interface measures the accuracy of a reported answer.

The record organizes the physical content of an explanation into the obligations in Table~\ref{tab:obligations}. An observation window specifies which fast processes must be retained, a transformation specifies which relations must be preserved, and an intervention specifies which response must change. Each successful test then supports a concrete part of the explanation. Alternative reductions are evaluated through the same declared quantities and tolerances.

\begin{table*}[t]
\centering\small
\caption{Physical commitments made explicit by the evaluation record. Each row connects a scientific claim to a quantitative test and the capability it supports.}
\label{tab:obligations}
\begin{tabularx}{\textwidth}{@{}P{.19\textwidth}YY@{}}
\toprule
Obligation & Test object & Supported physical capability\\
\midrule
Observable validity & Defined norm, scale, and observation window & Resolving displacement and initial velocity on their relevant time scales\\
Controlled reduction & Small parameter, retained variables, omitted terms & Locating the regime in which eliminated processes remain perturbative\\
Physical transformation & Covariance under the specified change & Preserving the physical relation under the specified transformation\\
Intervention response & A changed equation, coupling, or structural assignment & Predicting the response to a changed physical assignment\\
Mechanism discrimination & Information from allowed measurements at fixed cost & Selecting measurements that separate competing physical responses\\
\bottomrule
\end{tabularx}
\end{table*}

\subsection{A residual certificate with an explicit physical scope}

Residual-based error control and goal oriented estimates have a substantial numerical-analysis foundation\cite{cao2004adjoint}. Green function bounds also support validated computations for singularly perturbed boundary value problems\cite{birrell2015green}. We use the same residual-to-error principle in a known initial value problem, obtaining explicit constants for two observables and separating initial mismatch from fast residual layers. These constants make the approximation record directly evaluable.

Consider the dimensionless damped oscillator
\begin{equation}
 \eps x''+x'+x=0,\qquad x(0)=1,\quad x'(0)=0,
 \label{eq:oscillator}
\end{equation}
where $\eps=mk/\gamma^2$ and $\tau=kt/\gamma$ for mass $m$, stiffness $k$, damping $\gamma$, and physical time $t$. Primes denote derivatives with respect to $\tau$. For $0<\eps\le1/4$, the causal Green function $G_\eps$ of the differential operator is nonnegative and has integral one on $[0,\infty)$. Write $M_\eps=\sup_{\tau\ge0}G_\eps(\tau)$. For $\eps<1/4$,
\begin{align}
 a&=\frac{2}{1+\sqrt{1-4\eps}},\quad b=\frac{1+\sqrt{1-4\eps}}{2\eps},\nonumber\\
 M_\eps&=a\exp\left[-\frac{a\log(b/a)}{b-a}\right],
 \label{eq:kernelmax}
\end{align}
with $M_{1/4}=2/e$ by continuity.

\begin{proposition}[Displacement certificate]
\label{prop:residual}
For $0<\eps\le1/4$ and $T>0$, let $y\in C^2([0,T])$ be a candidate trajectory for Eq.~\eqref{eq:oscillator}, and let $R=\eps y''+y'+y=\sum_j R_j$. On $[0,T]$, with $e_0=y(0)-1$ and $e_1=y'(0)$,
\begin{align}
 \|y-x\|_\infty\le{}& |e_0|+\eps M_\eps|e_1|\nonumber\\
 &+\sum_j\min\{\|R_j\|_\infty,M_\eps\|R_j\|_1\}.
 \label{eq:certificate}
\end{align}
The norms of each residual component are taken over $[0,T]$.
\end{proposition}

The proof is given in Appendix~\ref{app:residual}. The bound separates initial data mismatch, persistent residuals, and short residual layers. A rapidly decaying residual can have a large instantaneous amplitude and a small integrated effect on displacement. The maximum norm controls persistent forcing, whereas the integrated norm captures a brief impulse. Choosing between them for each residual component retains the time scale of the eliminated process within an explicit displacement guarantee.

For the leading reduction $y_0=e^{-\tau}$, Eq.~\eqref{eq:certificate} gives
\begin{equation}
 b_0=\eps(1+M_\eps).
 \label{eq:leadingbound}
\end{equation}
A matched first order displacement is
\begin{equation}
 y_1=(1+\eps)e^{-(1+\eps)\tau}-\eps e^{-\tau/\eps}.
 \label{eq:matched}
\end{equation}
Its initial derivative mismatch is $-2\eps-\eps^2$. Splitting the slow and fast residuals yields
\begin{align}
 b_1={}&(2\eps+\eps^2)\eps M_\eps
 +\eps^2(1+\eps)(2+\eps)+M_\eps\eps^2.
 \label{eq:matchedbound}
\end{align}
Thus the certificate itself recovers $b_0=O(\eps)$ and $b_1=O(\eps^2)$ directly from the equation and the residual structure.

\begin{proposition}[Velocity certificate]
\label{prop:velocity}
Under the assumptions of Proposition~\ref{prop:residual},
\begin{align}
 \|y'-x'\|_\infty\le{}& M_\eps|e_0|+|e_1|\nonumber\\
 &+\sum_j\min\{2M_\eps\|R_j\|_\infty,\eps^{-1}\|R_j\|_1\}.
 \label{eq:velocitycertificate}
\end{align}
\end{proposition}
The derivative of the Green function weights the residual differently from the displacement kernel. For the two reductions, this gives
\begin{align}
 b_{v,0}&=1+2M_\eps\eps,\nonumber\\
 b_{v,1}&=3\eps+\eps^2
 +2M_\eps\eps^2(1+\eps)(2+\eps).
 \label{eq:velocitybounds}
\end{align}
The first remains order unity, while the second is $O(\eps)$. Appendix~\ref{app:residual} establishes the kernel identities and the bounds.

\subsection{Transformations and interventions require different responses}

For a physical transformation $g$, the observable obeys the corresponding map $Q(g\bm p)=T_gQ(\bm p)$. Unit changes, label permutations, and removable gauge phases belong in this class when their assumptions hold. An intervention changes a coupling, boundary condition, or structural assignment and specifies a corresponding response. We evaluate paired correctness alongside transformation consistency, so that the score records both the preserved relation and agreement with the physical reference.

For a specified intervention pair, we test
\begin{equation}
 e_{\rm int}=\frac{|\widehat{\Delta Q}-\Delta Q_{\mathcal M}|}{s_{\Delta Q}},
 \label{eq:interventionmetric}
\end{equation}
with a prespecified nonzero response scale. When observations admit several intervention responses, the record identifies the compatible mechanisms and an informative measurement that separates them\cite{pearl1995,causalrep,invariant}. In this way the explanation leads either to a verified response or to a concrete experimental question.

\section{Physical models and computational methods}
\label{sec:methods}

\subsection{Fast--slow dynamics}

The ratio $\eps$ in Eq.~\eqref{eq:oscillator} compares the inertial relaxation time $m/\gamma$ with the restoring time $\gamma/k$. We compare the exact solution, $y_0$, and $y_1$ over $\eps\in[10^{-4},2]$. A grid of 220 logarithmically spaced parameters and 5000 time points combining logarithmic and uniform sampling on $[0,8]$ resolves the initial layer. Velocity is evaluated over both $[0,8]$ and the delayed window $[0.1,8]$. Continuous displacement boundaries at tolerance 0.05 are located by optimizing over temporal error extrema and applying a bracketed root solver in $\eps$. Independent Radau integration verifies the exact trajectories, including critical damping. Numerical integration and root finding use SciPy\cite{scipy}.

\subsection{Effective exchange and learned representations}

We use the symmetric two-site Hubbard model with two electrons,
\begin{equation}
 H=-t\sum_\sigma(c^\dagger_{1\sigma}c_{2\sigma}+\mathrm{h.c.})
  +U\sum_{i=1}^2n_{i\uparrow}n_{i\downarrow},\quad U>0.
 \label{eq:hubbard}
\end{equation}
There is one undirected bond, $\hbar=1$, and the spins are $1/2$. The exact dimer provides a transparent reference for effective-model reasoning\cite{hubbarddimer}. Virtual high-energy excitations motivate a low-energy transformation of the type formalized by Schrieffer and Wolff\cite{schriefferwolff}. The triplet energy is zero and the relevant singlet block is
\begin{equation}
 H_S=\begin{pmatrix}0&-2t\\-2t&U\end{pmatrix}.
\end{equation}
Here h.c. denotes the Hermitian conjugate. The positive singlet--triplet gap is
\begin{equation}
 J=\frac{\sqrt{U^2+16t^2}-U}{2},\qquad r=|t|/U,
 \label{eq:gap}
\end{equation}
corresponding to $H_{\rm eff}=J(\bm S_1\cdot\bm S_2-1/4)$. We also evaluate $J_2=4t^2/U$ and $J_4=4t^2/U-16t^4/U^3$. Numerical evaluation uses the rationalized expression $J=8t^2/(\sqrt{U^2+16t^2}+U)$ to avoid cancellation.

We compare a raw multilayer perceptron (MLP) with a dimensionless multilayer perceptron ($\Pi$-MLP), where $\Pi$ denotes dimensionless groups. Both networks have two hidden layers of 32 hyperbolic tangent units and a linear output. They use the limited memory Broyden--Fletcher--Goldfarb--Shanno (L-BFGS) optimizer\cite{lbfgs}, implemented in scikit-learn\cite{sklearn}. The raw MLP maps $(\log(|t|/E_0),\log(U/E_0))$ to $\log(J/E_0)$, where $E_0$ is a fixed reference energy. The $\Pi$-MLP maps $\log r$ to $\log(J/U)$ and reconstructs $J$ by multiplication by $U$, imposing the dimensional homogeneity exactly. This comparison isolates the effect of a supplied physical representation within a common network architecture.

Each of 16 independent fits uses 240 physical training points sampled uniformly in logarithmic coordinates from $U/E_0\in[0.5,2]$ and $r\in[0.02,0.2]$. Each fit has 512 test points in each of three distributions comprising the training range, an energy scale shift with $U/E_0\in[20,200]$, and a coupling ratio shift with $r\in[0.3,1]$. One range changes at a time, and the two representations share the physical samples. Training uses exact gap labels, with test points reserved for evaluation. Inputs are standardized using training statistics, and the natural logarithmic targets retain their original scale. Besides relative gap error, we measure the response exponent $n_t=\partial\log J/\partial\log|t|$ at fixed $U$, using centered differences with logarithmic step $10^{-3}$. Central values are medians of the per fit test medians; percentile ranges describe variation across the repeated fits.

\subsection{A known observational and interventional model}

To isolate causal response, we use centered, dimensionless structural equations driven by independent standard normal exogenous noises\cite{pearl1995},
\begin{equation}
 P=\eta_P,\quad Z=P+0.1\eta_Z,\quad
 Y=2Z-3P+0.1\eta_Y,
 \label{eq:scm}
\end{equation}
Here $P$ is a measured common cause, $Z$ an intermediate quantity, and $Y$ the response. Each of 400 independent repetitions contains 400 observational samples and 400 randomized samples. The randomized assignment is $Z^{\rm do}\sim\mathcal N(0,1)$, independent of $P$ and both outcome noises. It produces $Y^{\rm do}=2Z^{\rm do}-3P+0.1\eta_Y^{\rm do}$, with a fresh independent standard-normal $\eta_Y^{\rm do}$. Observational and randomized conditions share the same 400 draws of $P$, but use independent $Z$-assignment and outcome noises. All ordinary least-squares regressions include an intercept; unadjusted and $P$-adjusted estimates use the same observational samples. The coordinates are dimensionless and have zero population means, with no distribution-specific rescaling.

\subsection{Two-mechanism measurement design}

Consider equal-prior hypotheses
\begin{equation}
 H_1:\ \mu_1(x)=x,\qquad H_2:\ \mu_2(x)=\frac{x}{1+x},
 \label{eq:hypotheses}
\end{equation}
with independent observations $y=\mu_H(x)+\xi$, $\xi\sim\mathcal N(0,\sigma^2)$, $\sigma=0.3$, and $x\in[0.02,1]$. Gains, saturation scale, and noise are known, and all queries have equal cost. Bayesian design selects a query by expected posterior information gain\cite{bayesdesign}. We use 40-point Gauss--Hermite quadrature on 48 logarithmically spaced candidate controls. Under these equal-variance known hypotheses, the best query is $x=1$ for every nondegenerate posterior, so the information strategy repeats that control. The known hypotheses give an analytically tractable reference for measurement selection.

We compare it with uniform controls on $[0.02,1]$, logarithmically uniform controls, and repeated $x=0.05$ measurements. The eight-query experiment has 500 noise repetitions, evenly divided between the two true hypotheses, with common noise sequences paired across policies. The outcome is the fraction with posterior probability at least 0.95 on the true hypothesis at each budget. Wilson intervals quantify finite repetition uncertainty\cite{wilson1927}. The comparison isolates measurement selection for these two known physical mechanisms.

\section{Results}
\label{sec:results}

\subsection{Initial layer matching improves displacement and velocity at different orders}

\begin{figure*}[t]
\centering
\includegraphics[width=\textwidth]{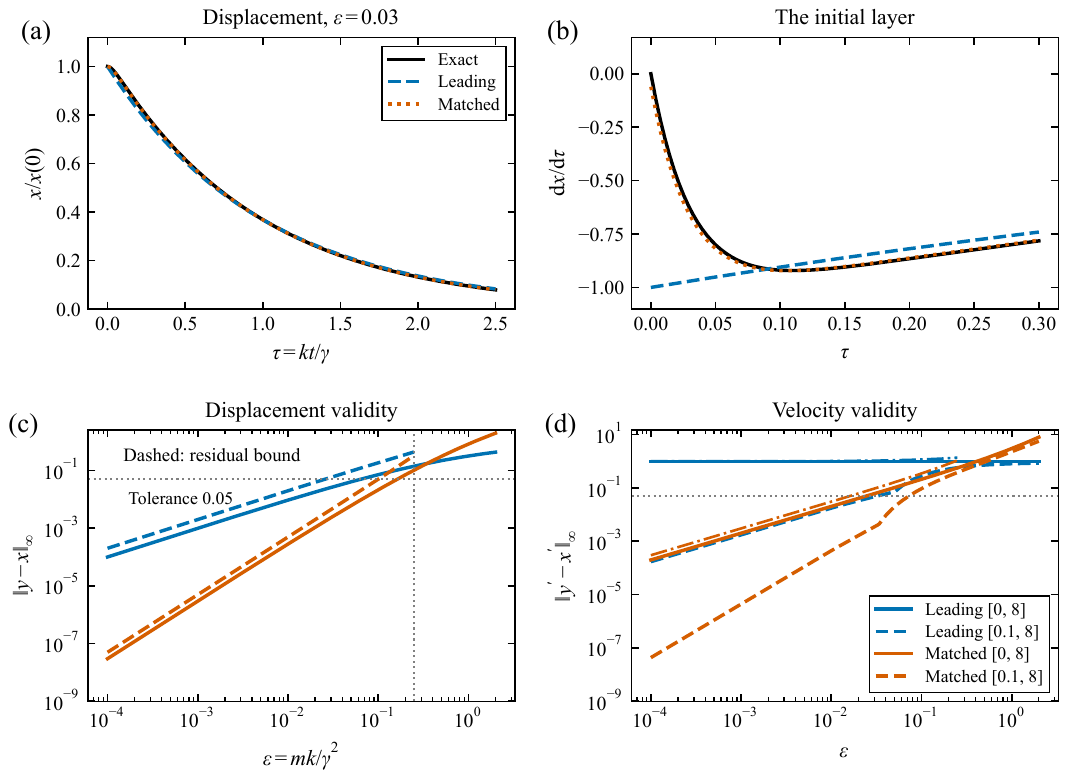}
\caption{Observable-dependent validity of the damped-oscillator reduction. (a) Displacement and (b) early velocity at $\eps=0.03$. (c) Solid displacement-error curves cover the full plotted parameter range; dashed residual certificates apply only for $\eps\le1/4$. The horizontal line is tolerance 0.05 and the vertical line marks the onset of oscillatory exact motion. (d) Velocity errors on full (solid) and delayed (dashed) windows, with full window certificates (dash-dot) restricted to $\eps\le1/4$. All errors are deterministic, using the initial displacement and dimensionless time unit as reference scales.}
\label{fig:observable}
\end{figure*}

As shown in Fig.~\ref{fig:observable}(a,b), the exact dynamics separate slow displacement from the initial velocity layer. At $\eps=0.03$, both approximate displacements are close to the exact trajectory. The leading solution nevertheless starts with $y_0'(0)=-1$, whereas the reference velocity is zero. For the full window,
\begin{equation}
 \|y_0'-x'\|_{[0,8],\infty}\ge1
 \label{eq:velocityfailure}
\end{equation}
for every positive $\eps$. This persistent initial mismatch identifies the fast momentum relaxation that must be restored for a full window velocity approximation. Panel (d) shows how a delayed observation window instead selects the slow dynamics.

The mismatch originates in eliminating momentum before its initial relaxation. The fast displacement component has amplitude $O(\eps)$ and duration $O(\eps)$, so its derivative contributes at order unity even as its displacement vanishes. Consequently, the zero-inertia and initial-time limits do not commute for velocity. The matched solution restores this component, leaving displacement and full window velocity errors of orders $\eps^2$ and $\eps$, respectively. Fitted exponents over $\eps\le0.003$ are 0.993 and 1.993 for the leading and matched displacement errors, and 0 and 1.000 for the respective full window velocity errors. The observed slopes confirm this separation of scales.

Figure~\ref{fig:observable}(c,d) connects this hierarchy to usable parameter domains. At displacement tolerance 0.05, the continuously located validity boundaries are $\eps=0.06498$ for the leading reduction and $0.16390$ for the matched reduction. Their sufficient certified boundaries are $0.02593$ and $0.10103$. Matching therefore enlarges both the directly accepted displacement interval and the analytically guaranteed interval. The matched velocity has its own sufficient boundary, $\eps=0.01624$, because differentiating the fast layer amplifies its contribution. The gap between each bound and the measured error quantifies the conservatism introduced by separately bounding residual components.

These accuracy boundaries all precede the critical value $\eps=1/4$, where the two real decay rates coalesce. Beyond this value, they form a complex conjugate pair. Consequently, an overdamped trajectory can already require a more accurate reduction before any oscillatory motion appears. The record distinguishes a physical change of motion from an observable dependent accuracy requirement. It also shows how the required measurement window determines the variables that an effective description must retain.

\subsection{Energy agreement and response agreement have different domains}

\begin{figure*}[t]
\centering
\includegraphics[width=\textwidth]{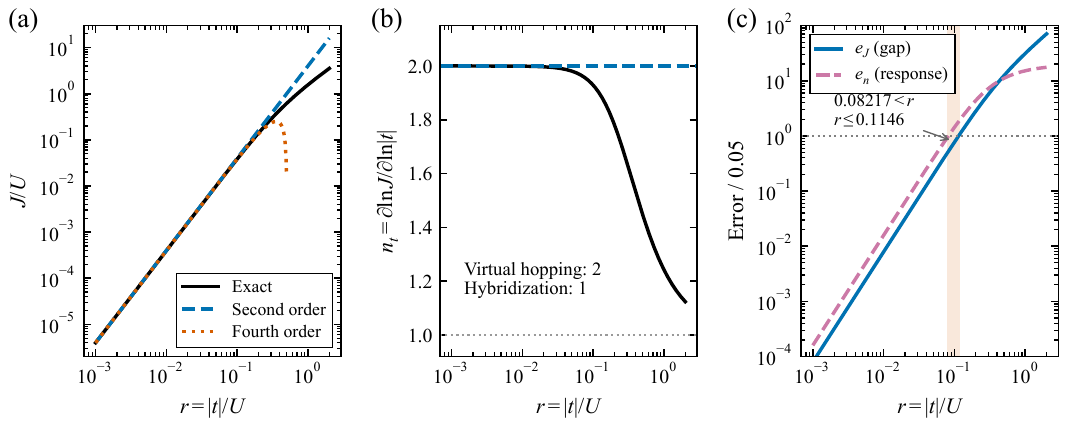}
\caption{Effective exchange and the scope of a superexchange explanation. (a) Exact gap and truncated expansions. The fourth order curve is shown only where positive; it becomes negative for $r>1/2$. (b) Exact hopping-response exponent. (c) Relative gap error and absolute exponent error of the second order reduction, each divided by its declared tolerance 0.05. The shaded interval $0.08217<r\le0.1146$ passes the gap criterion but fails the exponent criterion. Curves are exact analytical relations of the dimer model.}
\label{fig:exchange}
\end{figure*}

The dimer gap has the controlled expansion
\begin{equation}
 J=\frac{4t^2}{U}-\frac{16t^4}{U^3}+O(t^6/U^5).
 \label{eq:exchangeexpansion}
\end{equation}
The second order term reflects a virtual intermediate state with a doubly occupied site. Two hopping events connect the low-energy spin configurations, with an energy cost $U$ for the intermediate charge excitation. Singlet mixing lowers the energy, whereas Pauli exclusion prevents this lowering for the triplet, giving antiferromagnetic $J>0$. The exact response exponent is
\begin{equation}
 n_t=1+\frac{1}{\sqrt{1+16r^2}}.
 \label{eq:response}
\end{equation}
As illustrated in Fig.~\ref{fig:exchange}(b), $n_t$ decreases from 2 in the superexchange regime toward 1 as charge hybridization becomes strong. For small $r$, the two hopping events and their charge energy denominator give $J\propto t^2/U$. For large $r$, the exact singlet--triplet gap approaches $2|t|$, so the dominant energy dependence is linear in hopping. The exponent is a logarithmic susceptibility to hopping and tests the response dependence predicted by the proposed exchange mechanism.

For $J_2$, the exact relative gap error and absolute exponent error are
\begin{align}
 e_J&=\frac{\sqrt{1+16r^2}-1}{2},\nonumber\\
 e_n&=1-\frac{1}{\sqrt{1+16r^2}}.
 \label{eq:exchangeerrors}
\end{align}
Consequently, $e_J\le\delta$ permits
\begin{equation}
 r\le\frac{\sqrt{\delta(1+\delta)}}{2},
 \label{eq:gapdomain}
\end{equation}
whereas $e_n\le\eta<1$ requires
\begin{equation}
 r\le\frac14\sqrt{(1-\eta)^{-2}-1}.
 \label{eq:responsedomain}
\end{equation}
At $\delta=\eta=0.05$, the respective limits are 0.1146 and 0.08217. The shaded interval in Fig.~\ref{fig:exchange}(c) therefore isolates points accepted by the gap criterion that require a more accurate response description. The tolerances represent 5\% relative gap error and 0.05 absolute exponent error. Expanding Eq.~\eqref{eq:exchangeerrors} gives $e_J=4r^2+O(r^4)$ and $e_n=8r^2+O(r^4)$, explaining why response is the more sensitive diagnostic near the controlled limit. A small correction to the energy can contribute more strongly to its logarithmic derivative.

The fourth order curve in Fig.~\ref{fig:exchange}(a) further identifies the role of expansion control. The Taylor series has radius of convergence $1/4$ in $r$, and its fourth order truncation becomes negative at $r>1/2$ while the exact gap remains positive. In the coupling shift test its median relative error is about 130\%, compared with 69.6\% at second order. These analytical controls provide a physical interpretation of the crossover results. Resumming the hybridization or using the exact reduced function is appropriate when the expansion parameter reaches this regime.

The completed records in Table~\ref{tab:workedrecords} translate these curves into individual decisions. At $\eps=0.05$, direct verification accepts the leading displacement while the residual bound supplies a more conservative guaranteed domain. At $r=0.1$, the same second order expression supports the gap value under its tolerance and calls for refinement of the hopping response. Recording these outcomes separately identifies both the usable prediction and the physical quantity that benefits from an improved approximation.

\begin{table*}[t]
\centering\small
\caption{Filled evaluation records. All listed acceptance tolerances are 0.05. The oscillator uses $[0,8]$; the dimer quantities are static, with the derivative at fixed $U$.}
\label{tab:workedrecords}
\begin{tabularx}{\textwidth}{@{}P{.19\textwidth}YY@{}}
\toprule
Record field & Leading oscillator, $\eps=0.05$ & Second-order dimer, $r=0.1$\\
\midrule
Targets and scales & Displacement, $s_x=1$ & Gap $s_J=J$; exponent $s_n=1$\\
Actual errors & $e_x=0.04008$ & $e_J=0.03852$; $e_n=0.07152$\\
Bound or reference & $b_0=0.09452$ & Exact error expressions, Eq.~\eqref{eq:exchangeerrors}\\
Validity decisions & Displacement passes & Gap passes; response fails\\
Certificate decision & Not guaranteed by this bound & Exact expressions resolve both decisions\\
\bottomrule
\end{tabularx}
\end{table*}

\subsection{Homogeneity protects energy scale transfer}

\begin{figure*}[t]
\centering
\includegraphics[width=.98\textwidth]{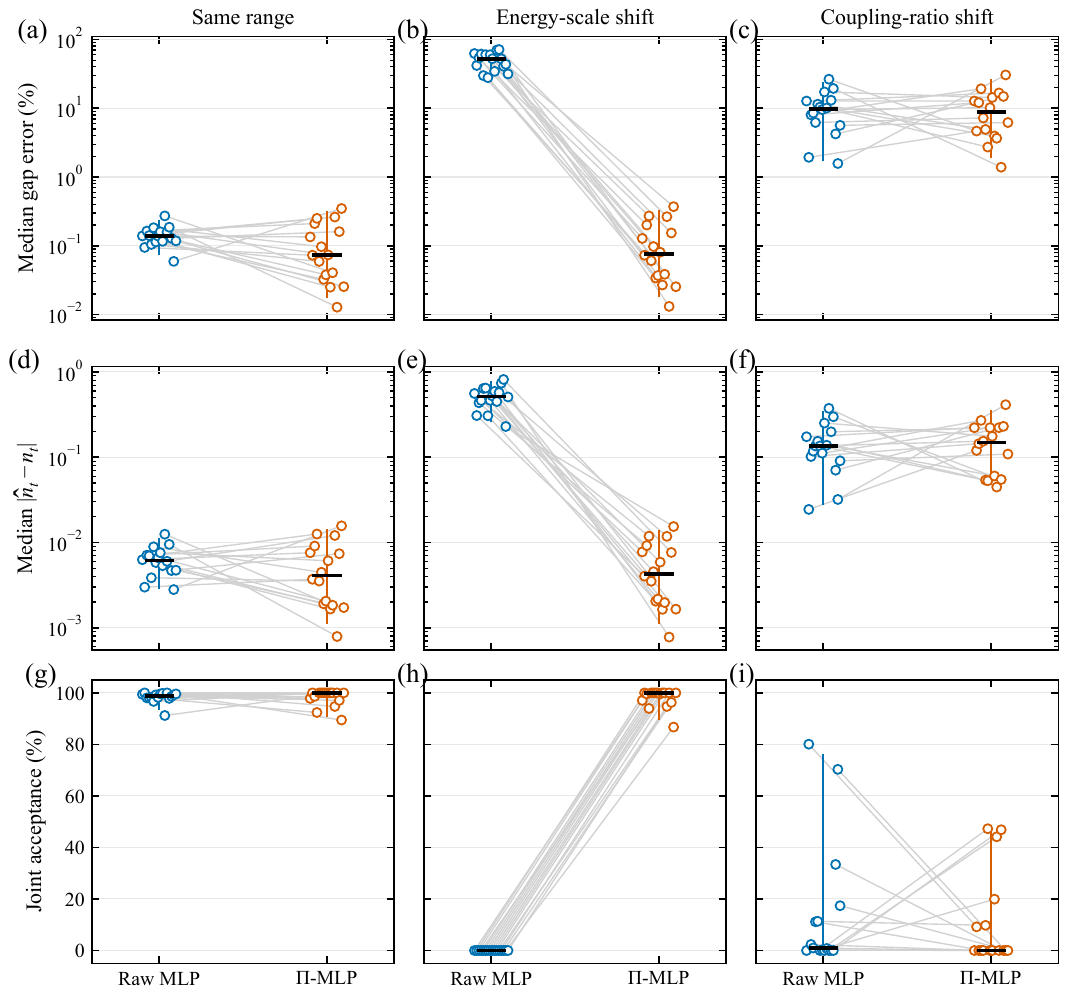}
\caption{Gap errors (top), hopping-response exponent errors (middle), and joint acceptance fractions (bottom) in 16 repeated network fits. A bottom-row point is the fraction of 512 test points simultaneously satisfying relative gap error $\le0.05$ and absolute exponent error $\le0.05$; other points are per fit error medians. Gray segments connect fits using the same physical samples. Black bars are medians across fits and colored vertical lines span their 2.5--97.5 percentiles, describing training and sampling variation.}
\label{fig:neural}
\end{figure*}

The exact relation $J(\lambda t,\lambda U)=\lambda J(t,U)$, for $\lambda>0$, identifies the energy scale transformation. It leaves $r$ and the competition between hopping and repulsion unchanged. Table~\ref{tab:neural} and Fig.~\ref{fig:neural} show how enforcing that relation changes the learned model. The raw network has a median same range gap error of 0.139\%, but its error rises to 52.3\% when the overall energy scale changes. The dimensionless network has errors of 0.0733\% and 0.0768\%, respectively. The improvement follows from a supplied physical prior: scale transfer requires no extrapolation of the dimensionless function when $r$ remains in its training range.

\begin{table*}[t]
\centering\small
\caption{Neural fits evaluated against the exact dimer. Each value is the median across 16 independent fits of a 512-point test median. Gap errors are percentages; response-exponent errors are absolute dimensionless differences.}
\label{tab:neural}
\begin{tabular}{llrr}
\toprule
Distribution & Representation & Gap error (\%) & Exponent error\\
\midrule
Same range & Raw MLP & 0.1388 & 0.006155\\
Same range & $\Pi$-MLP & 0.07332 & 0.004088\\
Energy-scale shift & Raw MLP & 52.33 & 0.5163\\
Energy-scale shift & $\Pi$-MLP & 0.07675 & 0.004289\\
Coupling-ratio shift & Raw MLP & 9.813 & 0.1356\\
Coupling-ratio shift & $\Pi$-MLP & 8.71 & 0.1492\\
\bottomrule
\end{tabular}
\end{table*}

The middle row of Fig.~\ref{fig:neural} shows that the protected transfer extends to a physical derivative. The raw network's median exponent error rises from 0.00616 to 0.516, whereas the dimensionless network changes from 0.00409 to 0.00429. Writing $\widehat J=U\widehat f(r)$ makes the reason explicit. Its hopping exponent is $r\widehat f'(r)/\widehat f(r)$, independent of the overall energy scale. The same representation thus preserves the relation between energy and response along each scaling orbit. The Euler homogeneity identity provides a complementary check of this imposed structure.

Changing $r$ probes the shape of the crossover function. In $r\in[0.3,1]$, gap errors are 8.71\% for the dimensionless network and 9.81\% for the raw network, with corresponding exponent errors of 0.149 and 0.136. These results locate the additional learning problem within $\widehat f(r)$ itself. In this range the exact exponent moves away from the quadratic superexchange limit toward linear hybridization, so data that resolve the crossover provide information beyond the scaling relation. Appendix~\ref{app:homogeneity} makes this distinction precise through a family of functions sharing the same homogeneity, training values, and limiting laws while varying within an unobserved interval.

Joint acceptance makes this distinction directly operational. For each trained fit and the declared test distribution, define
\begin{align}
 p_J&=\Pr(e_J\le0.05),\quad p_n=\Pr(e_n\le0.05),\nonumber\\
 p_{\cap}&=\Pr(e_J\le0.05\ \land\ e_n\le0.05),
 \label{eq:jointacceptance}
\end{align}
where probabilities are empirical fractions in each fit's test set, summarized by their median across fits. As reported in Table~\ref{tab:joint}, the median joint fraction under energy scale transfer rises from zero for the raw representation to 100\% for the dimensionless representation, whose 2.5--97.5 percentile range is 89.4--100\%. The bottom row of Fig.~\ref{fig:neural} displays the paired changes directly. Under coupling ratio transfer, the dimensionless model has a median gap acceptance of 30.7\% and joint acceptance of zero. The derivative criterion identifies a concrete target for additional training, namely the change in the physical response through hybridization.

Joint acceptance also gives a practical interpretation to the error summaries. A median prediction error describes a typical point, whereas $p_{\cap}$ estimates the probability mass, under the declared test distribution, on which both physical requirements are satisfied. Reporting $p_J$, $p_n$, and $p_{\cap}$ together therefore connects surrogate accuracy to its usable domain. The paired fits in the figure show how strongly this domain changes when the representation incorporates the correct scale relation.

\begin{table*}[t]
\centering\small
\caption{Median acceptance fractions for the two physical tolerances in Eq.~\eqref{eq:jointacceptance}. Each fit uses 512 test points shared by the two representations. Variation across fits is shown in Fig.~\ref{fig:neural}.}
\label{tab:joint}
\begin{tabular}{llrrr}
\toprule
Distribution & Representation & $p_J$ (\%) & $p_n$ (\%) & $p_{\cap}$ (\%)\\
\midrule
Same range & Raw MLP & 100 & 98.9 & 98.9\\
Same range & $\Pi$-MLP & 100 & 100 & 100\\
Energy-scale shift & Raw MLP & 1.17 & 0 & 0\\
Energy-scale shift & $\Pi$-MLP & 100 & 100 & 100\\
Coupling-ratio shift & Raw MLP & 20.8 & 0.879 & 0.879\\
Coupling-ratio shift & $\Pi$-MLP & 30.7 & 0 & 0\\
\bottomrule
\end{tabular}
\end{table*}

We additionally transform each of the 8192 same range test points per representation by $\lambda\in\{0.01,0.1,10,100,1000\}$, keeping $r$ fixed. The maximum dimensionless-network defect $|\widehat J(\lambda t,\lambda U)/(\lambda\widehat J(t,U))-1|$ is $2.45\times10^{-15}$ across all five transformations. At $\lambda=100$, the raw-network defect has a pooled median of 0.622. These strictly paired tests isolate covariance from the independent test sampling used in Table~\ref{tab:neural}, explaining why the two dimensionless gap-error summaries can differ slightly while the covariance relation is exact to rounding precision.

\subsection{Causal structure recovers the intervention response}

For Eq.~\eqref{eq:scm}, $\operatorname{Var}(Z)=1.01$, $\operatorname{Var}(Y)=1.05$, and $\operatorname{Cov}(Z,Y)=-0.98$. The unadjusted population slope and explained variance are
\begin{align}
 \beta_{\rm obs}&=2-3/1.01=-0.9703,\nonumber\\
 R^2&=\frac{0.98^2}{1.01\times1.05}=0.9056.
 \label{eq:observational}
\end{align}
An intervention on $Z$ instead has
\begin{equation}
 \frac{\partial\E[Y\mid\doop(Z=z)]}{\partial z}=2.
 \label{eq:doslope}
\end{equation}
In the observational data, an increase of $Z$ is usually accompanied by an increase of $P$. The negative direct effect of $P$ on $Y$ exceeds the positive $Z$ pathway, producing the negative association. Randomization replaces the assignment of $Z$ and cuts the link from $P$, while preserving the response equation for $Y$. It therefore isolates the positive response coefficient $+2$.

\begin{figure*}[t]
\centering
\includegraphics[width=.96\textwidth]{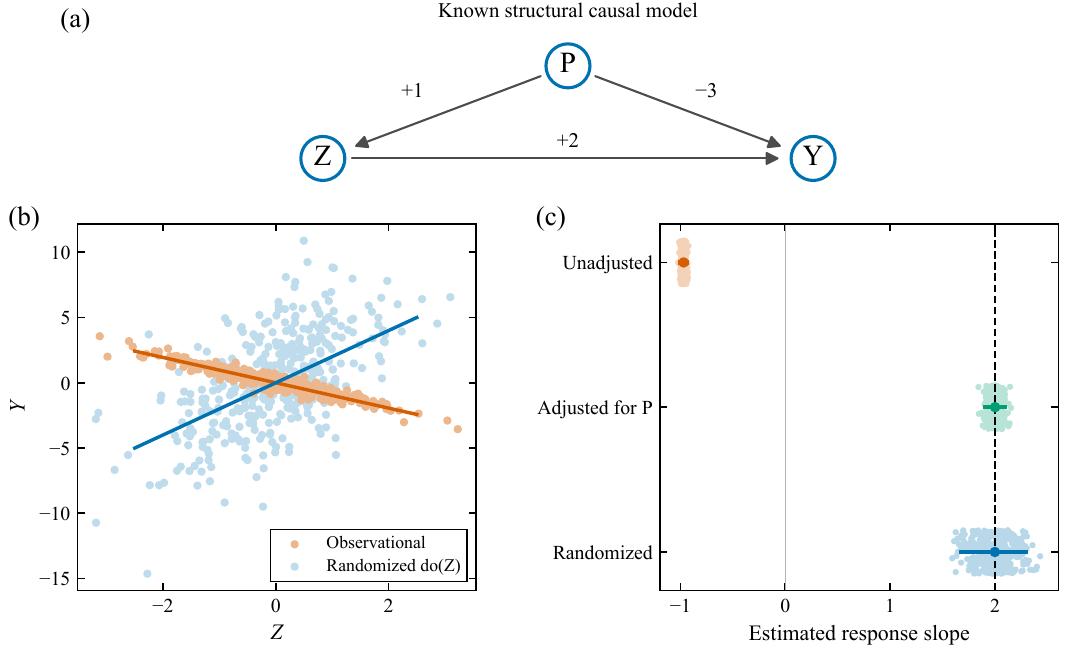}
\caption{Observational and intervention response in a known synthetic system. (a) Structural graph and coefficients. (b) One repetition with 400 samples per condition and population regression lines; randomized $Z^{\rm do}\sim\mathcal N(0,1)$ is independent of $P$. (c) Slopes over 400 repetitions; points are medians, segments are 2.5--97.5 percentiles, and the dashed vertical line is the true intervention slope. Variables use the dimensionless coordinates of the structural equations, with no distribution-specific rescaling.}
\label{fig:causal}
\end{figure*}

Figure~\ref{fig:causal}(c) confirms the population analysis over 400 repetitions. The observational regression has median $R^2=0.9046$ and mean slope $-0.9699$ with standard deviation 0.0160. Adjustment for the measured common cause recovers a mean slope of 2.0017 with standard deviation 0.0504, while randomization gives 1.9941 with standard deviation 0.1532. For a fixed exogenous state and $\Delta z=2$, the structural response is $\Delta Y=4$, compared with approximately $-1.94$ from observational regression. Specifying the intervention thus changes both the sign and the physical meaning of the predicted response.

The scatter and slope distributions clarify why precision and causal accuracy are distinct. The observational slope is tightly concentrated around the association generated by the common cause. Adjustment uses the measured $P$ to separate its direct contribution from the $Z$ response, while the unadjusted randomized regression averages over variation in $P$. This explains the smaller spread of the adjusted estimator in this supplied structural model. Both procedures recover the positive response through their respective identification conditions.

An identifiability test specifies which additional observation would determine the response. With independent standard Gaussian noises and $0<|\rho|<1$, the two models
\begin{align}
 Z=\eta_Z,&\quad Y=\rho Z+\sqrt{1-\rho^2}\eta_Y,\nonumber\\
 Y=\eta_Y,&\quad Z=\rho Y+\sqrt{1-\rho^2}\eta_Z
 \label{eq:equivalent}
\end{align}
generate the same observational distribution and have $Z$-intervention slopes $\rho$ and zero, respectively. Temporal information, a justified physical direction, or an intervention can separate these mechanisms. Reporting this equivalence class together with a discriminating measurement turns observational ambiguity into an actionable experimental prediction.

\subsection{Mechanism separation sets the measurement cost}

For two known Gaussian hypotheses with mean difference $\Delta(x)=|\mu_1(x)-\mu_2(x)|$ and equal variance $\sigma^2$, a single measurement at fixed control has Kullback--Leibler (KL) divergence
\begin{equation}
 \KL(H_1\Vert H_2\mid x)=\frac{\Delta(x)^2}{2\sigma^2}.
 \label{eq:querykl}
\end{equation}
After $B$ independent measurements at that control, the log posterior odds in favor of the true hypothesis are Gaussian with mean $B\Delta^2/(2\sigma^2)$ and variance $B\Delta^2/\sigma^2$. Equal prior odds are assumed.

\begin{proposition}[Fixed-control query requirement]
\label{prop:budget}
For $0<\alpha,\beta<1/2$ and $\Delta>0$, the smallest integer budget ensuring posterior probability at least $1-\alpha$ on the true hypothesis with probability at least $1-\beta$ is
\begin{align}
 B_{\min}(x)=\left\lceil\frac{\sigma^2}{\Delta(x)^2}
 \left[z_{1-\beta}+\sqrt{z_{1-\beta}^2+2L_\alpha}\right]^2\right\rceil,
 \label{eq:budget}
\end{align}
where $L_\alpha=\log[(1-\alpha)/\alpha]$ and $z_{1-\beta}=\Phi^{-1}(1-\beta)$.
\end{proposition}

Appendix~\ref{app:budget} derives this expression. If the first difference between mechanisms is $\Delta(x)=c x^k+o(x^k)$ with $c>0$ and $k>0$, then at fixed $\sigma$, $\alpha$, and $\beta$,
\begin{equation}
 B_{\min}(x)=\Theta(x^{-2k})\quad(x\to0).
 \label{eq:infoscaling}
\end{equation}
The exponent $2k$ connects the first distinguishing physical correction to the measurement cost for the specified Gaussian hypotheses and fixed control.

\begin{figure*}[t]
\centering
\includegraphics[width=\textwidth]{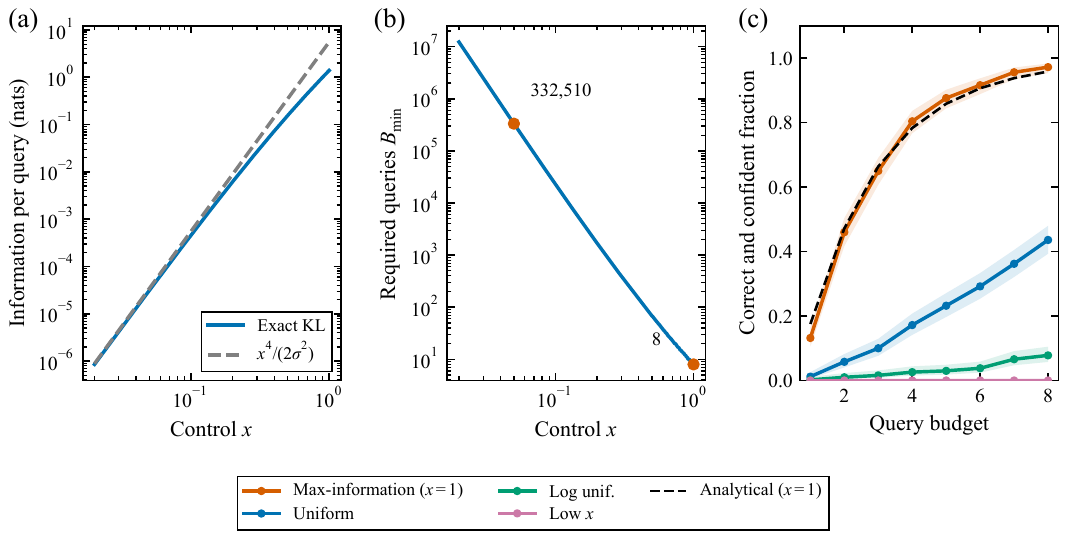}
\caption{Mechanism separation and query cost. (a) Exact per-query KL divergence and its small-control fourth order asymptote. (b) Integer query requirement from Eq.~\eqref{eq:budget} for $\alpha=\beta=0.05$, at noise standard deviation 0.3. (c) Equal-budget measurement strategies over 500 balanced noise repetitions; shaded regions are pointwise Wilson 95\% intervals. The black dashed curve is the analytical success probability at $x=1$ from Eq.~\eqref{eq:confidenceprobability}. The known hypotheses make repeated $x=1$ measurement the maximum-information policy.}
\label{fig:information}
\end{figure*}

For Eq.~\eqref{eq:hypotheses}, $\Delta=x^2/(1+x)$ and $k=2$. Both mechanisms share the linear response, with their first difference arising from the quadratic saturation correction. As shown in Fig.~\ref{fig:information}(a,b), this physical degeneracy produces an $x^4$ information scale and an $x^{-4}$ discrimination cost. With $\alpha=\beta=0.05$ and $\sigma=0.3$, Eq.~\eqref{eq:budget} requires eight measurements at $x=1$ and 332,510 at $x=0.05$. Moving the control into the nonlinear regime separates the mechanisms relative to the same observation noise.

The policy comparison in Fig.~\ref{fig:information}(c) gives the corresponding finite budget result. Eight measurements produce correct and sufficiently confident fractions of 97.2\% for maximum information selection, 43.6\% for uniform controls, and 7.8\% for controls sampled uniformly in logarithmic coordinates. Repeated low control measurements remain below the posterior threshold. The exact success probability at $x=1$ rises from 17.53\% at one query to 95.84\% at eight; the simulated eight query fraction differs by 1.52 Monte Carlo standard errors. The analytical curve therefore provides a quantitative check across the full budget range.

The ordering of these strategies follows their access to the distinguishing saturation term. Uniform sampling allocates some measurements near the informative upper control, whereas logarithmic sampling concentrates more strongly where the shared linear response dominates. The maximum information policy uses the largest allowed separation at every measurement. In this known two mechanism problem, the result gives a direct physical reason for the gain from experimental design\cite{bayesdesign}. It also identifies a useful evaluation target for a scientific agent, namely selecting a control that reveals the relevant correction within a fixed measurement budget.

\section{Implications for evaluating language models}
\label{sec:evaluation}

The controlled examples define evaluation targets for approximation validity, transformation consistency, intervention response, and measurement selection. The development interface implements structured questions about prescribed reductions and physical models, with numerical and categorical outputs scored against analytical references.

The evaluation set contains 128 base systems and 256 paired prompts in four families. Oscillator tasks return actual displacement validity and certificate status separately; dimer tasks return distinct gap and response decisions. The distinction in Eq.~\eqref{eq:domains} thus becomes a scored scientific output. Field correctness, strict correctness, paired joint correctness, and equivalence consistency quantify complementary aspects of the responses. Both variants of each system are presented independently and remain in the same data split. Missing or unparseable responses remain in the score denominators.

These questions support comparisons between structured physical prompts and general analysis prompts under the same response format and tool budget. Supplying the governing equations isolates the use of physical structure in validity and response calculations. A further evaluation condition can ask a model to construct equations from scenarios and data, following the control of prior information studied in PhysGym\cite{physgym}. Separating these conditions identifies the contributions of model construction and reasoning within a supplied model.

Interactive evaluations can extend these tasks by allowing a model to choose successive measurements and tracking the information gained within a fixed query budget. The Gaussian calculation supplies an analytical reference for measurement selection with known candidate mechanisms. Analogous extensions can ask a model to propose a reduction and then test its accepted observables and domain. Appendix~\ref{app:interface} defines the current task families, data splits, scoring rules, and representative analytical answers. The task bank and scoring implementation are available from the corresponding author on request.

\section{Discussion}

Validity changes with the observable and time window because eliminating a fast variable affects derivatives more strongly than slowly varying coordinates. In the oscillator, the leading displacement converges uniformly as $O(\eps)$ while the initial velocity mismatch remains unity. Matching the initial layer improves displacement to $O(\eps^2)$ and velocity to $O(\eps)$, whereas delaying the observation window targets the response after fast relaxation. Propositions~\ref{prop:residual} and \ref{prop:velocity} express this distinction through the Green function norms appropriate to each observable. The dimer gives a corresponding static example. Near the exchange limit, the relative gap error is $4r^2+O(r^4)$ and the absolute response-exponent error is $8r^2+O(r^4)$, producing different acceptance boundaries at the same numerical tolerance. These calculations identify which variables and corrections must be retained for a specified physical measurement.

The neural comparison separates energy scale transfer from changes in coupling. Writing the predicted gap as $U$ times a learned function of $r$ enforces homogeneity and preserves the response exponent along a scaling orbit. Under the energy scale shift, the dimensionless network has a median joint acceptance fraction of 100\%. Under the coupling ratio shift, the corresponding fraction is zero, although the median gap acceptance fraction is 30.7\%. The second test therefore identifies where additional training information is needed, namely the shape and derivative of the crossover function. As shown by the function family in Appendix~\ref{app:homogeneity}, crossover observations determine information independent of homogeneity and the limiting laws. New data in that range can be selected to improve both the gap and its hopping response.

Mechanism discrimination depends on the measured variables, allowed interventions, and observation noise. In the causal example, measuring the common cause or randomizing the intermediate variable recovers the positive response; the observationally equivalent models in Eq.~\eqref{eq:equivalent} specify an intervention that distinguishes their responses. In the measurement example, the candidate mechanisms share a linear response and first separate through the quadratic saturation term. Their per-query information consequently scales as $x^4$, giving an $x^{-4}$ query requirement at fixed noise. Increasing the control from 0.05 to 1 reduces the required budget from 332,510 to eight measurements for the stated confidence targets. The gain follows from exposing the distinguishing nonlinear correction. Specifying the accessible control range, noise, and candidate responses thus connects a proposed mechanism to a quantitative measurement decision.

\section{Conclusion}

Validity records for individual observables make the scope of a physical explanation quantitatively testable. We linked reductions to declared scales and tolerances, separated certified domains from actual validity, and derived complementary displacement and velocity certificates. Exact dimer relations and paired neural tests show why energy homogeneity protects scale transfer while joint acceptance of gap and response identifies the coupling crossover as a separate learning target. A known causal model separates observational association from intervention response. The order at which two mechanisms separate then determines an exact fixed control query cost under Gaussian noise.

These results support evaluation that keeps prediction, approximation, and mechanism identification individually accountable. The analytical references and structured development interface connect each validity decision to a physical quantity, an acceptance criterion, and a reproducible calculation, providing a basis for evaluating physical reasoning in scientific AI.

\section*{Data and code availability}
The data, code, trained model weights, task bank, and scoring implementation supporting this study are available from the corresponding author on request.

\section*{AI assistance disclosure}
Generative AI assisted literature discovery, analytical development, code implementation, figure preparation, and manuscript drafting. Analytical expressions, explicit structural equations, and independently checked numerical calculations define the reference results. The authors take responsibility for the analysis, numerical results, references, and final manuscript.

\FloatBarrier
\appendix

\section{Proof and scope of the residual certificate}
\label{app:residual}

For $\eps<1/4$, the Green function is
\begin{equation}
 G_\eps(\tau)=\frac{e^{-a\tau}-e^{-b\tau}}{\eps(b-a)},\quad\tau\ge0,
\end{equation}
where $a,b$ are defined in Eq.~\eqref{eq:kernelmax} and $ab=1/\eps$. It is nonnegative and its integral is one. Its maximum occurs at $\tau_*=\log(b/a)/(b-a)$, giving Eq.~\eqref{eq:kernelmax}. At critical damping, $G_{1/4}=4\tau e^{-2\tau}$ and the same statements hold by continuity.

Let $\phi_0$ be the exact solution with initial displacement one and initial velocity zero. It decreases from one to zero in this parameter range. The error satisfies
\begin{equation}
 y-x=e_0\phi_0+e_1\eps G_\eps+\sum_j G_\eps*R_j.
\end{equation}
The two convolution bounds $\|G*R_j\|_\infty\le\|R_j\|_\infty$ and $\|G*R_j\|_\infty\le M_\eps\|R_j\|_1$ establish Proposition~\ref{prop:residual}. The nonnegative kernel on $0<\eps\le1/4$ establishes the stated scope of this certificate.

For $y_1$, the residual separates as
\begin{align}
 R_s&=\eps^2(1+\eps)(2+\eps)e^{-(1+\eps)\tau},\nonumber\\
 R_f&=-\eps e^{-\tau/\eps}.
\end{align}
Bounding $R_s$ in maximum norm and $R_f$ in integrated norm, together with the initial velocity error, gives Eq.~\eqref{eq:matchedbound}. The fast residual has amplitude $O(\eps)$ but integrated magnitude $O(\eps^2)$, explaining the importance of matching the residual norm to the eliminated process.

For velocity, $\phi_0'=-G_\eps$ and differentiation of the convolution gives
\begin{equation}
 y'-x'= -e_0G_\eps+e_1\eps G_\eps'+\sum_jG_\eps'*R_j.
\end{equation}
The kernel is unimodal and vanishes at both endpoints. The following kernel norms are taken over $[0,\infty)$; residual norms remain over $[0,T]$.
\begin{equation}
 \|G_\eps'\|_1=2M_\eps,\qquad
 \|G_\eps'\|_\infty=\eps^{-1}.
 \label{eq:kernelderivative}
\end{equation}
The positive derivative maximum is $G_\eps'(0)=1/\eps$. Its negative extremum occurs at $2\log(b/a)/(b-a)$ and has magnitude smaller than $1/\eps$; the critical case follows by continuity. Young's convolution inequalities now establish Proposition~\ref{prop:velocity}. For $y_1$, use the maximum norm for $R_s$ and the integrated norm for $R_f$. Their contributions are $2M_\eps\eps^2(1+\eps)(2+\eps)$ and $\eps$, respectively. Together with $|e_1|=2\eps+\eps^2$, these give Eq.~\eqref{eq:velocitybounds}.

\section{Independent information in the crossover function}
\label{app:homogeneity}

Write $J(t,U)=Uf(r)$, $r=|t|/U$, with $f(r)>0$. Let a smooth nonnegative bump $h(r)$ have compact support in a test interval disjoint from the training interval. Then
\begin{equation}
 \widetilde J(t,U)=U[f(r)+c\,h(r)],\qquad c>0,
\end{equation}
has the same homogeneity, hopping sign invariance on the single bond, and training values as $J$. Since the bump vanishes near zero and infinity, both asymptotic regimes also remain unchanged. Varying $c$ changes the value and derivative within its support while preserving these constraints. This family explicitly isolates the information supplied by crossover observations from that supplied by symmetry, training values, and limiting laws.

\section{Derivation of the Gaussian query requirement}
\label{app:budget}

At a fixed $x$, the log likelihood ratio from one observation is
\begin{equation}
 \ell=\frac{(y-\mu_2)^2-(y-\mu_1)^2}{2\sigma^2}.
\end{equation}
Under $H_1$, it is normal with mean $\Delta^2/(2\sigma^2)$ and variance $\Delta^2/\sigma^2$; under $H_2$, reverse its sign. The distributions of evidence in favor of the true hypothesis are identical. Summing $B$ independent observations and using equal priors gives
\begin{align}
 \Prob\{p(H_{\rm true}\mid D_B)\ge1-\alpha\}
 =\Phi\left(\frac{\sqrt B\Delta}{2\sigma}-\frac{\sigma L_\alpha}{\sqrt B\Delta}\right).
 \label{eq:confidenceprobability}
\end{align}
Set the argument at least $z_{1-\beta}$ and solve the resulting quadratic in $\sqrt B\Delta/\sigma$ to obtain Eq.~\eqref{eq:budget}. The result quantifies the sampling probability of attaining the declared posterior threshold on the true mechanism.

\section{Training and uncertainty}
\label{app:reproducibility}

Both networks use double precision and natural logarithms. Input centering and scaling are fitted on the training samples, and logarithmic targets retain their original scale. Training uses an $L_2$ penalty of $10^{-5}$, a gradient tolerance of $10^{-9}$, and at most 2000 iterations. Each repeated fit varies the physical samples and initialization, with both representations using the same training and test points. This pairing compares the two representations under matched physical conditions.

The oscillator and dimer curves describe deterministic changes of physical parameters. The network percentiles describe training and sampling variation across repeated fits. Causal intervals summarize finite samples from the stated structural model, and measurement intervals summarize independent observation noise under the two hypotheses. Independent simulation with 50,000 balanced noise sequences verifies the analytical success curve over budgets 1--8, with maximum probability discrepancy 0.00389. These uncertainty descriptions give each plotted range a specific statistical interpretation.

\section{Executable development interface}
\label{app:interface}

Each of four families has 32 base systems divided into 16 training, eight development, and eight test groups. Across families this gives 64, 32, and 32 groups, corresponding to 128, 64, and 64 prompts. Paired variants remain in the same split. Oscillator variants rescale all equation coefficients, and dimer variants rescale energies and reverse the hopping sign on the single bond. Causal and measurement variants test consistency under equivalent wording and parameter ordering. The oscillator is stratified into certified points, accepted points with a bound above tolerance, rejected points with a bound above tolerance, and points outside the proof domain. The dimer is stratified by joint gap and response acceptance, gap acceptance alone, and rejection of both. Each split contains every realizable stratum. For each sampled parameter $p$, the boundary margin satisfies
\begin{equation}
 \min_{b\in\mathcal B}\frac{|p-b|}{b}\ge10^{-3},
 \label{eq:boundarymargin}
\end{equation}
where $\mathcal B$ contains the relevant positive boundaries. For the oscillator, $p=\eps$ and the boundaries are the leading displacement certificate threshold, the actual displacement threshold on $[0,8]$, and $1/4$. For the dimer, $p=r$ and the boundaries are the gap and response acceptance thresholds. The generator uses unrounded boundary values; the sampling configuration specifies ranges, logarithmic sampling, boundary adjacent points, and randomization.

\begin{table*}[t]
\centering\small
\caption{Stratified coverage of the oscillator and dimer task families, counted by base system. Each system contributes two equivalent prompts.}
\label{tab:coverage}
\begin{tabular}{llrrr}
\toprule
Family & Physical region & Training & Development & Test\\
\midrule
Oscillator & Certified & 4 & 2 & 2\\
Oscillator & Valid, bound above tolerance & 4 & 2 & 2\\
Oscillator & Invalid, bound above tolerance & 4 & 2 & 2\\
Oscillator & Outside proof domain & 4 & 2 & 2\\
Dimer & Gap and response accepted & 6 & 3 & 2\\
Dimer & Gap accepted, response rejected & 5 & 3 & 3\\
Dimer & Gap and response rejected & 5 & 2 & 3\\
\bottomrule
\end{tabular}

\end{table*}

As summarized in Table~\ref{tab:coverage}, 16 oscillator prompts distinguish actual acceptance from a sufficient guarantee, and 22 dimer prompts distinguish energy acceptance from response acceptance. These cases make the two central validity decisions directly testable within every data split.

Let task $i$ have $m_i$ required fields and $c_{if}\in\{0,1\}$ indicate a correct field. Let $a_i=\prod_f c_{if}$ be strict task correctness, and let $N=256$, $G=128$. The evaluation scores are defined as
\begin{align}
 S_{\rm field}&=\frac{\sum_{i,f}c_{if}}{\sum_i m_i},\quad
 S_{\rm strict}=\frac1N\sum_i a_i,\nonumber\\
 S_{\rm joint}&=\frac1G\sum_{g=1}^G a_{g0}a_{g1},\quad
 S_{\rm equiv}=\frac1G\sum_{g=1}^G q_g,
 \label{eq:graderscores}
\end{align}
where $q_g=1$ when all paired answers agree after mapping to the common representation. The first score weights the 1280 required fields equally, the second requires every field of a task to be correct, and the third requires both variants of a system to be correct. The fourth measures equivalence consistency after validating field types and permitted certificate states. Reporting the last two together separates preserved relations from agreement with the analytical reference. Field scores are also grouped into validity decisions (256 fields), numerical inference (384), qualitative regime identification (64), and structural rules (576), giving each physical obligation an explicit contribution to the evaluation.

A numerical answer $a$ matches reference $b$ when $|a-b|\le10^{-5}+0.01|b|$. Paired equivalence uses the symmetric rule $|a-b|\le10^{-5}+0.01\max(|a|,|b|)$. Boolean answers use strict true or false values. Certificate status distinguishes a sufficient guarantee, a bound above tolerance, and parameters outside the proof domain. Missing answers, invalid types, and nonfinite numbers receive zero for the corresponding fields, and all tasks remain in the score denominators. Different derivations are accepted when they produce the requested quantities and decisions. Explanatory reasoning and proposed alternative reductions can be assessed alongside these structured outputs.

\subsection*{Representative physical tasks}

\noindent\textit{Oscillator.} For $m=0.055222103$, $\gamma=1.2826983$, $k=1.4897249$, evaluate the leading and matched reductions on $[0,8]$ at tolerance 0.05. The reference gives $\eps=0.05$, nonoscillatory motion, and displacement error powers 1 and 2. The leading displacement is accepted with $e_x=0.04008$, while its bound $b_0=0.09452$ is above tolerance.

\noindent\textit{Hubbard dimer.} For $t=0.077362401$, $U=0.77362401$, the reference gives $r=0.1$, $J/U=0.038516481$, leading hopping power 2, and invariance under hopping sign reversal. The second order gap error $e_J=0.03852$ satisfies tolerance 0.05; its response error $e_n=0.07152$ requires refinement.

\noindent\textit{Causal response.} In $Z=aP+\sigma_Z\eta_Z$ and $Y=bZ+cP+0.1\eta_Y$, take independent standard normal exogenous variables and $a=1.1302027$, $b=1.3579449$, $c=-2.8713394$, $\sigma_Z=0.15652024$. The observational slope is -1.1347991 and the intervention slope is 1.3579449. Adjustment for measured $P$ recovers the positive response.

\noindent\textit{Measurement design.} Compare $y=x$ and $y=x/(1+x/s)$ with equal priors and $s=1.837815$, $\sigma=0.37099509$, $x\in[0,1.1757526]$. The maximum information control is the upper endpoint. The saturated response lies below the linear response for $x>0$, and their difference vanishes as $x\to0$. At $x=0$ the likelihoods coincide and the posterior odds remain unchanged.

These analytical answers satisfy all required fields and receive full field and strict correctness scores. Paired variants yield the same answers after the specified physical mapping.

\FloatBarrier
\printbibliography

\end{document}